\documentclass{article} 
\usepackage{iclr2026_re-align_workshop,times}

\usepackage{amsmath,amsfonts,bm}

\def\eqref#1{equation~\ref{#1}}

\def\1{\bm{1}}

\DeclareMathAlphabet{\mathsfit}{\encodingdefault}{\sfdefault}{m}{sl}
\SetMathAlphabet{\mathsfit}{bold}{\encodingdefault}{\sfdefault}{bx}{n}

\DeclareMathOperator*{\argmin}{arg\,min}

\usepackage{hyperref}
\usepackage{url}

\title{\textbf{Atlas-Alignment}: Making Interpretability Transferable Across Language Models}

\author{
\textbf{Bruno Puri\textsuperscript{1,2}},
\textbf{Jim Berend\textsuperscript{1}},
\textbf{Sebastian Lapuschkin\textsuperscript{1,3}},
\textbf{Wojciech Samek\textsuperscript{1,2,4}}
\\
\\
\textsuperscript{1}Department of Artificial Intelligence, Fraunhofer Heinrich Hertz Institute\\
\textsuperscript{2}Department of Electrical Engineering and Computer Science, Technische Universität Berlin\\
\textsuperscript{3}Centre of eXplainable Artificial Intelligence, Technological University Dublin
\\
\textsuperscript{4}BIFOLD - Berlin Institute for the Foundations of Learning and Data\\
\\
\small{\textbf{corresponding authors:} \texttt{\{bruno.puri, wojciech.samek\}@hhi.fraunhofer.de}}
}

\usepackage{graphicx}
\usepackage{array}
\usepackage{multirow}
\usepackage{tabularx}
\usepackage{wrapfig}
\usepackage{xcolor}
\usepackage{booktabs}
\usepackage{siunitx}
\usepackage{listings}
\usepackage{mdframed}
\usepackage{xurl}
\usepackage{fvextra}

\newmdenv[
  linewidth=0.5pt,
  skipabove=\baselineskip,
  skipbelow=\baselineskip
]{promptbox}

\newcolumntype{C}[1]{>{\centering\arraybackslash}m{#1}}

\makeatletter
\newcommand{\shortto}{\mathrel{\mathpalette\shortto@{.65}}} 
\newcommand{\shortto@}[2]{\vcenter{\hbox{\scalebox{#2}[1]{$#1\to$}}}}
\newcommand{\Tsc}{\mathbf{T}_{s \shortto c}}
\makeatother

\iclrfinalcopy 

\begin{document}

\maketitle

\begin{abstract}
Interpretability is crucial for building safe, reliable, and controllable language models, yet existing interpretability pipelines remain costly and difficult to scale. Interpreting a new model typically requires training model-specific components (e.g., sparse autoencoders), followed by manual or semi-automated labeling and validation, imposing a growing “transparency tax” that does not scale with the pace of model development.
We introduce \textbf{Atlas-Alignment}, a framework that avoids this cost by aligning the latent space of a new model to a pre-existing, labeled Concept Atlas using only shared inputs and lightweight representational alignment methods. 
Through quantitative and qualitative evaluations, we show that simple alignment methods enable robust semantic retrieval and steerable generation without the need for labeled concept datasets. \textbf{Atlas-Alignment} thus amortizes the cost of explainable AI and mechanistic interpretability: by investing in a single high-quality Concept Atlas, we can make many new models transparent and controllable at minimal marginal cost.
\end{abstract}

\section{Introduction} \label{sec:introduction}

Large language models (LLMs) are increasingly deployed in domains where safety, reliability, and controllability are critical. Yet, their internal representations and processes remain largely opaque to their users, hindering verifiability and trust. Model activations capture the semantic and functional structure of processing, but without the tools to interpret them, we cannot rigorously assess how a model arrives at its outputs or intervene when it is behaving in unforeseen ways. Interpretability is thus essential for both trust and reliability, as well as for practical control of model behavior.
Advances in mechanistic interpretability and explainable AI have begun to uncover these latent structures using sparse autoencoders (SAEs) \citep{bricken2023monosemanticity} and automated feature-discovery pipelines \citep{bills2023language, choi2024automatic, dreyer2025mechanisticunderstandingvalidationlarge}. These can, for instance, extract latent features that are both monosemantic and can be described in natural language terms. Such features enable analysis of reasoning processes and make controlled interventions on model activations easier. However, current interpretability methods remain costly and difficult to scale. Each new model and layer requires training SAEs, generating feature descriptions, and validating them individually. The need to explain each new model variant from scratch makes comprehensive interpretability computationally expensive and often infeasible. 
\\
In this work, we pursue a complementary direction: rather than interpreting each model in isolation, we make use of work in representational alignment, to transfer interpretability across models. We introduce \textbf{Atlas-Alignment}, a framework for aligning the latent space of an unknown ``subject model'' to a well-understood, human-labeled latent space that we refer to as Concept Atlas. Once aligned, the subject model inherits the interpretability of the Concept Atlas: its features can be semantically queried, compared, and steered without the need for costly SAE training or labeled concept datasets.
\textbf{Atlas-Alignment} builds on two complementary hypotheses. The Linear Representation Hypothesis suggests that semantic concepts are often linearly encoded as directions in latent spaces \citep{park2024linearrepresentationhypothesisgeometry}, while the Platonic Representation Hypothesis suggests that different LLMs converge on broadly similar latent structures \citep{huh2024platonicrepresentationhypothesis}. Together, these imply that a single, carefully constructed Concept Atlas could serve as a universal ``concept hub''. By aligning subject models to this atlas using only shared input data and lightweight transformations, we can recover semantic structure and enable plug-and-play interpretability across a wide range of models.
This translation unlocks several capabilities. Aligned models support semantic search and grouping of features, cross-model and cross-layer comparison of representations, and controllable steering of generation along human-interpretable directions, allowing us to navigate yet unexplored latent spaces --- all without training probes or SAEs, or generating synthetic datasets. Crucially, the cost of interpretability is amortized: a single high-quality Concept Atlas can make many new models transparent and steerable at a minimal marginal cost.
Our contributions are as follows:%
\begin{enumerate}%
    \item We introduce \textbf{Atlas-Alignment}, a lightweight and general framework for transferring interpretability across language models using Concept Atlases, shared input data, and representational alignment methods, bypassing the need for model-specific SAE training.
    \item We benchmark multiple representational alignment methods and establish that Orthogonal Procrustes is sufficient to preserve semantic geometry, outperforming other methods in retrieval and steering.
    \item We demonstrate that models can ``inherit'' complex semantic controls from a Concept Atlas, enabling precise steering and feature search in opaque models without requiring any concept-specific labeled datasets.
\end{enumerate}%
In the following sections, we first review related work on interpretability and representational alignment (Sec.~\ref{sec:related-work}). We then introduce the \textbf{Atlas-Alignment} framework (Sec.~\ref{sec:method}). Next, we present qualitative and quantitative results on feature identification, semantic transfer, and concept steering (Sec.~\ref{sec:results}). Finally, we discuss implications and directions for future work (Sec.~\ref{sec:conclusion}).

\begin{figure*}[t!]
    \centering
    \includegraphics[width=1.\textwidth]{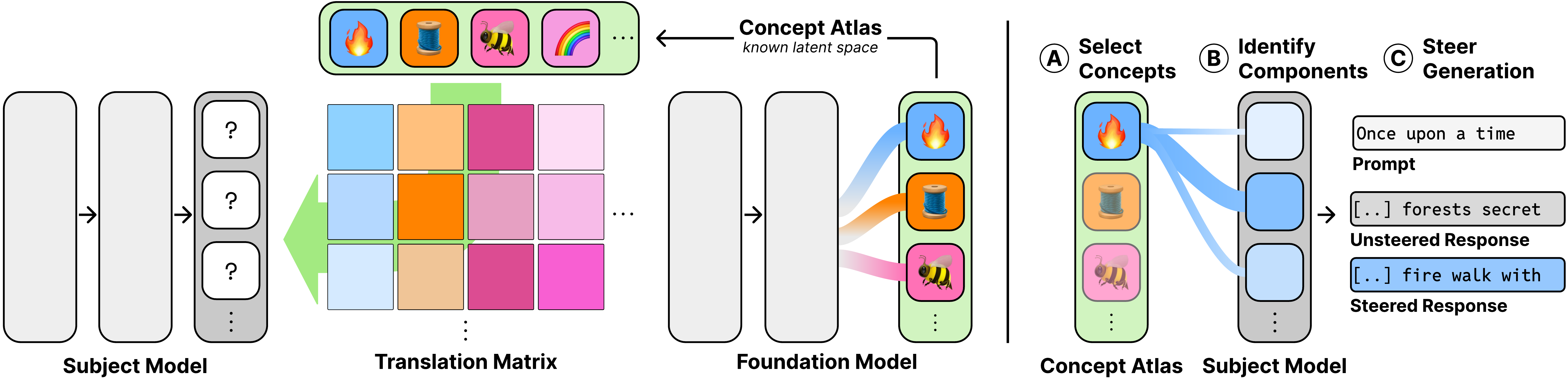}
    \caption{%
    \textbf{Atlas-Alignment} makes the latent space of a subject model interpretable by aligning it with a Concept Atlas --- a human-interpretable, labeled latent space. \textbf{Left}: The subject model’s hidden representations are mapped into the Concept Atlas, allowing each subject feature to be described as a linear combination of atlas concepts. \textbf{Right}: Once aligned, the method enables a range of interpretability tasks. (A) One or multiple concepts are selected from the Atlas, (B) corresponding subject model components are identified, or (C) the subject model’s output is steered along the concept direction.%
    }
    \label{fig:latent-space-alignment}
\end{figure*}

\section{Related Work} \label{sec:related-work}

\textbf{Features and Concepts} Neural network neurons and their linear combinations have been shown to encode human-aligned concepts to a surprising degree \citep{Achtibat2023, bykov2023dora}. Supervised methods for identifying such features typically rely on curated concept datasets to locate directions or neurons of interest \citep{kim2018interpretabilityfeatureattributionquantitative, bau2017networkdissectionquantifyinginterpretability}. Although effective, these methods are inherently limited to the pre-defined choice of concepts. 
Unsupervised approaches instead examine activations from large, unlabeled datasets to discover semantic clusters. A particularly influential line of work has been the development of sparse autoencoders (SAEs) \citep{bricken2023monosemanticity}, which decompose polysemantic features into sparse, more monosemantic and often highly interpretable features.
This progress has increased the need for methods that assign meaning to features in a scalable manner. Recent work automates the generation of natural language descriptions of features \citep{bills2023language, paulo2024automaticallyinterpretingmillionsfeatures, templeton2024scaling}, along with automated evaluation methods to assess their quality \citep{bills2023language, DBLP:conf/nips/KopfBHLHB24, puri-etal-2025-fade, gurarieh2025enhancingautomatedinterpretabilityoutputcentric}. However, these pipelines remain costly: each new latent space typically requires training SAEs, producing feature descriptions, and validating them before meaningful semantic interaction becomes possible, typically requiring the processing of billions of tokens.

\textbf{Aligned Latent Representations} A broad line of work studies how latent representations from different neural networks can be aligned or mapped into a common space. The Platonic Representation Hypothesis \cite{huh2024platonicrepresentationhypothesis} posits that sufficiently scaled models converge to shared latent structures across architectures and modalities, while the linear representation hypothesis \cite{park2024linearrepresentationhypothesisgeometry} suggests that many human-aligned concepts are encoded approximately linearly in latent space. Together, these hypotheses motivate the use of lightweight mappings that preserve semantic structure across models.
Building on this view, several works have approached cross-model alignment from a practical perspective. \citet{jha2025harnessinguniversalgeometryembeddings} train mappings using an unsupervised cycle consistency loss, while \citet{thasarathan2025universalsparseautoencodersinterpretable} propose training SAEs that decompose multiple models into a shared concept space. However, both approaches require expensive training. Other works more similar to ours show how simple linear transformations can suffice for cross-model transfer: \citet{maiorca2023latent} introduce a component-stitching approach to reuse components from other models to improve performance on downstream tasks such as classification or reconstruction. \citet{huang2025crossmodeltransferabilitylargelanguage} transfer supervised steering vectors via linear mappings learned from small contrastive datasets. 
In the vision domain, representational alignment has been applied specifically for interpretability by aligning internal representations to CLIP embeddings \cite{radford2021learningtransferablevisualmodels}, which allow mapping textual queries into a shared image–text embedding space. This semantic reference space is then used to assign meaning to internal features of vision models, enabling text-based querying, comparison, and validation of learned representations \cite{oikarinen2023clipdissectautomaticdescriptionneuron, Moayeri2023TextToConceptB, dreyer2025mechanisticunderstandingvalidationlarge}. While these works demonstrate that alignment to semantically meaningful spaces can support interpretability, they fundamentally rely on the CLIP model acting as a direct bridge between abstract concepts and internal representations. In language models, an analogous reference space is not readily available without relying on per-concept synthetic data generation.
\\
\\
Motivated by this gap, we propose \textbf{Atlas-Alignment}, which aligns an unknown model’s internal representations to a Concept Atlas: a sparse, more monosemantic, and labeled latent space, derived from an SAE. This enables the transfer of mechanistic interpretability, allowing semantic search and dataset-free controllable steering in new, opaque models without retraining or relabeling interpretability tools such as SAEs, reducing the cost of interpreting new models by one to two orders of magnitude.
\section{Method} \label{sec:method}

\begin{figure*}[t!]
    \centering
    \includegraphics[width=1.\textwidth]{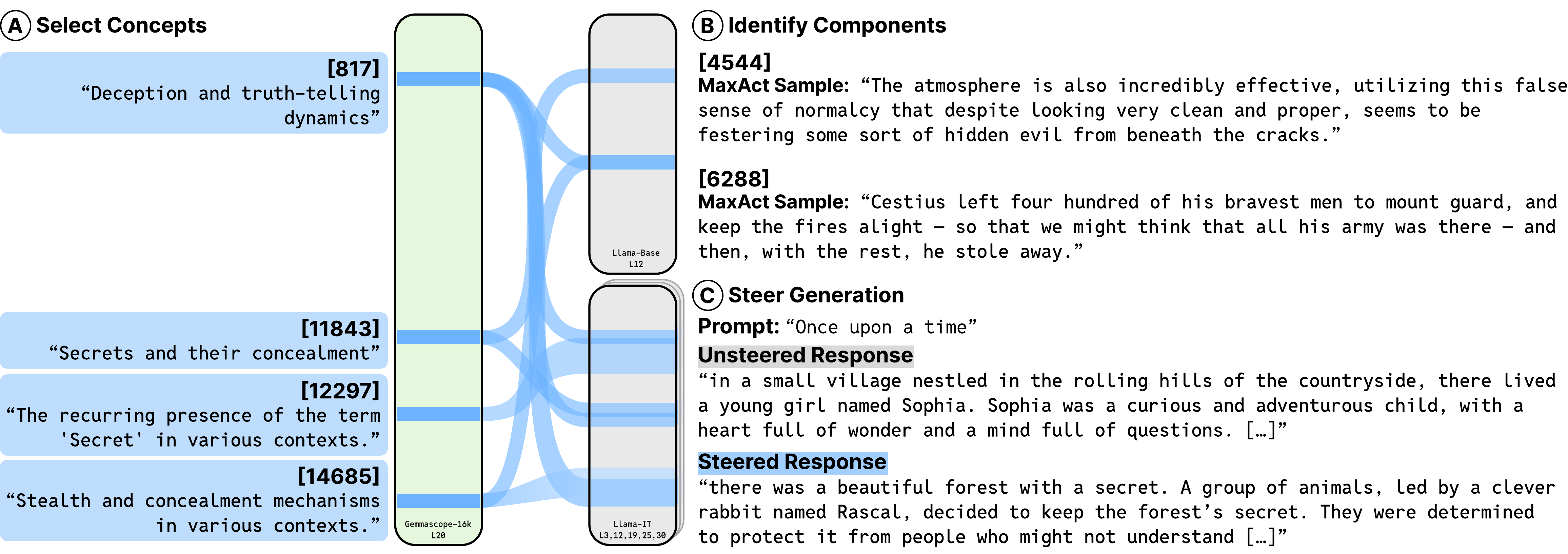}
    \caption{Examples of using \textbf{Atlas-Alignment} for identification and steering. (A) A Concept Query is constructed from multiple Concept Atlas features related to the theme of ``secrets and deception'' and mapped into the latent spaces of two subject models. (B) In Llama-Base, the alignment reveals two features in layer 12 that encode relevant concepts. (C) In Llama-IT, the same Concept Query is used to steer generation across multiple layers, shifting outputs toward concept-related text.}
    \label{fig:search-steer-method}
\end{figure*}

Our goal is to align the latent representations of a subject model (of which we have no prior understanding) with a Concept Atlas, a labeled and interpretable latent space derived from a foundation model. Once aligned, the subject model inherits the interpretability of the atlas: its features can be queried, compared, and modified along semantically meaningful directions.
This alignment requires only shared input data: by passing the same dataset through both models and comparing activations, we can construct lightweight mappings that reveal the semantic structure of otherwise uninterpretable features.

\subsection{Background and Notation} \label{sec:method-background}

Let $\mathbf{X} = \{x_1, \ldots, x_N\}$ be a dataset, where each sample $x_i$ is a sequence of text. The subject model $f_{s}: \mathcal{X} \rightarrow \mathcal{H}_s$, maps from data domain $\mathcal{X}$ into an intermediate feature space $\mathcal{H}_s$, an unknown space we aim to interpret. The foundation model $f_{f}: \mathcal{X} \rightarrow \mathcal{C}$ maps from the data domain into our Concept Atlas $\mathcal{C}$, a feature space that is semantically interpretable. Forwarding the dataset through the models and applying max-pooling over the sequence lengths, we retrieve aggregated activation matrices {\small$A_s \in \mathbb{R}^{N \times d_s}$ }and {\small $A_c \in \mathbb{R}^{N \times d_c}$}. To address differences in tokenization across models, we apply pooling over sequences, using max-pooling to retain the strongest per-sequence activations.

\subsection{Concept Atlas} \label{sec:method-concept-atlas}

The Concept Atlas is our reference space of interpretable features, where each dimension corresponds to a human-understandable concept. We rely on SAEs to construct our Concept Atlas due to their strength in producing sparse, monosemantic and human-interpretable latent spaces. Each Concept Atlas feature $c_k \in \mathcal{C}$ can be assigned a natural language description $d_k \in \mathcal{D}$ via manual or automatic labeling methods. This annotated atlas serves as a ``semantic dictionary'', where every feature corresponds to a concept humans can name and reason about. We can combine multiple features and weight them according to the concepts we want to identify or steer. Aligning new models to this space lets us carry over those semantics without repeating the costly process of building and labeling SAEs from scratch.

\subsection{Translating Latent Spaces} \label{sec:method-translation}

We define a translation function that expresses subject model features of a layer in terms of the Concept Atlas features%
{\small
\[
\begin{aligned}
\operatorname{translate}: \mathbb{R}^{N \times d_s} \times \mathbb{R}^{N \times d_c} \rightarrow \mathbb{R}^{d_s \times d_c}, (A_s, A_c) \mapsto \Tsc
\end{aligned}
\]
}%
Each row of the resulting matrix $\Tsc$ represents a subject model feature in terms of the Concept Atlas features. This function can be instantiated with standard representational alignment methods. As an example, we show the Orthogonal Procrustes method. All methods used in this work are described in Appendix \ref{sec:appendix-method-translation-methods}.

\textbf{Orthogonal Procrustes Translation:} Constrains the translation matrix to be orthogonal, restricting the alignment to a rotation or reflection of the space. %
{\small
\[
\begin{aligned}
\operatorname{translate}_{\text{ OP}}(A_s, A_c) &= \underset{\Tsc}{\arg\min} \ \| A_s - A_c \Tsc^\top \|_F^2 \\ \text{s.t.} \quad &\Tsc \Tsc ^\top = \mathbf{I}
\end{aligned}
\]
}%
With row-wise $L_2$ normalization, this is equivalent to minimizing cosine distance between activations.

\subsection{Using Latent Space Translations} \label{sec:method-using-translation}

Once the mapping is learned, we can use the matrix $\Tsc$ and the knowledge encoded in the Concept Atlas to make the subject model latent space both interpretable and controllable. This involves three steps: Creating a Concept Query, mapping it into the subject model latent space, and applying it for retrieval or steering. For a visualization of the approach, see Figure~\ref{fig:latent-space-alignment}. See Figure \ref{fig:search-steer-method} for an example.

\paragraph{1. Creating a Concept Query}

The \emph{Concept Query} $q_{c}\!\in\!\mathbb{R}^{d_c}$ is a vector that represents the concept of interest in terms of Concept Atlas features. We can construct it in several ways: Firstly, we can directly use the feature descriptions $\mathcal{D}$ of the Concept Atlas and set the indices of features that are relevant to the concept to a value of one, while setting the rest to zero. Secondly, we can use embedding models to embed both a human query and the feature descriptions $\mathcal{D}$ and retrieve the descriptions that are closest to it. We set the indices of relevant features to one (or a similarity score), and the rest to zero. Finally, similar to how one would create a steering vector, we can also build averaged or contrastive concept queries in the Concept Atlas, by forwarding a set of concept related sequences through the foundation model and aggregating their latent space activations. This flexibility allows us to use both human-guided and data-driven concept definitions.

\paragraph{2. Mapping to the subject model}
We map the concept query vector to the subject model space using the row-normalized cosine similarity between the query vector $q_c$ and the $\Tsc$ matrix 
{\small
 \[
s_c = 
\widehat{\mathbf{T}}_{s \shortto c} \frac{q_c}{\|q_c\rVert_2} \quad \text{where} \quad \widehat{\mathbf{T}}_{s \shortto c \ [k,:]} 
= 
\frac{\mathbf{T}_{s \shortto c \ [k,:]}}{\lVert \mathbf{T}_{s \shortto c \ [k,:]}\rVert_2}
\]
}%
The resulting similarity vector $s_{c}\!\in\!\mathbb{R}^{d_s}$ scores each subject feature by its alignment with the chosen concept.

\paragraph{3. Identification and Steering}

We use the similarity vector $s_c$ to \emph{identify} which subject model features encode the target concept. Each entry of $s_c$ measures how strongly a feature aligns with the concept. Investigating the top-scoring features answers the question: ``where, and to what degree, is the concept represented in the subject model’s latent space?'' In this way, the previously opaque feature space becomes searchable and interpretable.

The same vector $s_c$ also provides a direction for intervention. By adding a scaled version of $s_c$ to the subject model activations at inference time %
{\small
 \[
a^{\text{(modified)}}
= \Bigl(a^{\text{(original)}} + \lambda\,s_{c}\Bigr)
\cdot \frac{\lVert a^{\text{(original)}} \rVert_2}
{\lVert a^{\text{(original)}} + \lambda\,s_{c} \rVert_2}
\]
}%
we can steer the model’s behavior along the chosen concept, analogous to steering with vectors derived from labeled datasets. Here $\lambda$ is a scalar that controls the strength of the intervention. Because $s_c$ is obtained without supervision, this provides a fast, concept-data free mechanism to control model behavior.

\section{Experimental Results} \label{sec:results}

We first outline implementation details (Sec.~\ref{sec:results-details}). We then present qualitative results on feature identification and concept steering (Sec.~\ref{sec:results-qualitative}). Next, we report quantitative evaluations of semantic translation (Sec.~\ref{sec:results-evaluating-translation}) and steering effectiveness (Sec.~\ref{sec:results-evaluating-steering}).

\begin{figure*}[t!]
    \centering
    \includegraphics[width=\linewidth]{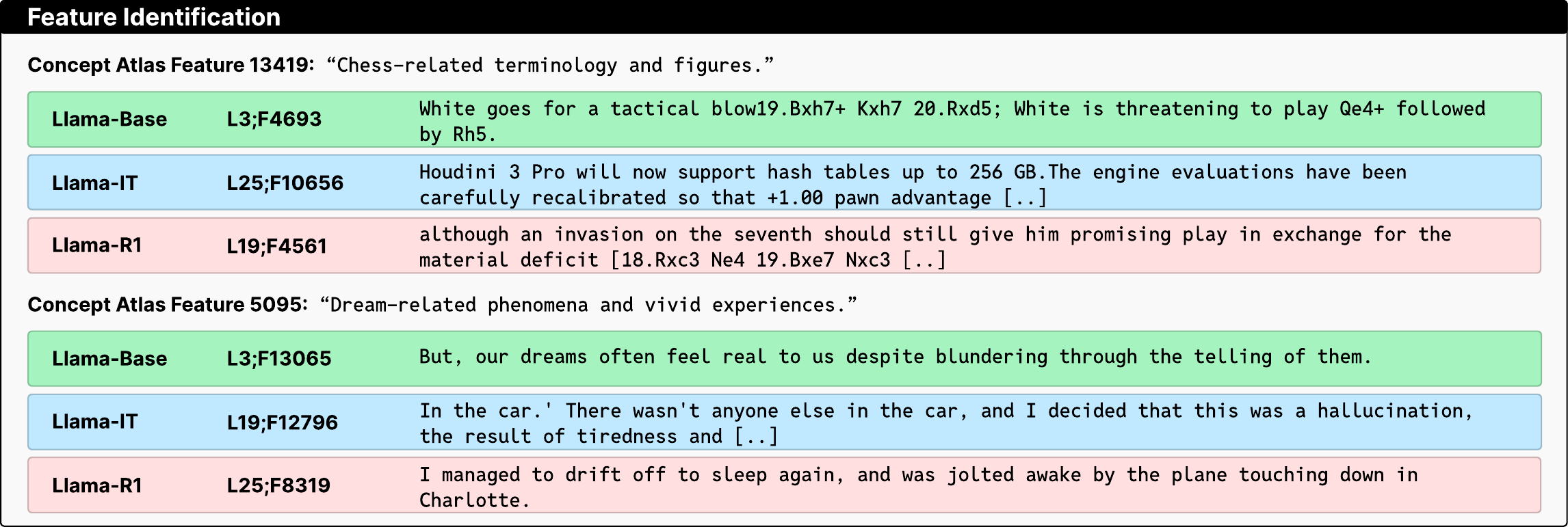}
    \caption{Cross-model feature identification via \textbf{Atlas-Alignment}. Concept queries from Gemma Scope 16k are mapped into Llama variants using Orthogonal Procrustes. The top-aligned features activate on semantically related samples, indicating successful semantic transfer, while highlighting how specialized neurons in the subject models can be detected.}
    \label{fig:search-qualitative-results}
\end{figure*}

\subsection{Implementation Details} \label{sec:results-details}

We use a subset of the Pile dataset \citep{gao2020pile}, preprocessed as described in Appendix~\ref{sec:appendix:dataset-generation}, comprising 1 million sequences and over 30 million tokens. We evaluate five translation methods: covariance, correlation, linear regression, Orthogonal Procrustes with row-wise $L_2$-normalization, and Semantic Lens. Each method is trained on 500k samples, with a disjoint set of 500k samples reserved for the qualitative and quantitative evaluations.
We consider subject model latent spaces from four model families: GPT-2 \citep{radford2019language} at layer 8; Qwen~2.5 \citep{qwen2, qwen2.5} in the 0.5B, 1.5B, and 3B configurations at layers 15, 17, and 21 respectively; Ministral-8B-Instruct-2410 \citep{ministral8b2024} at layer 21; and Llama~3.1~8B base, instruction-tuned, and R1-distilled variants \citep{grattafiori2024llama3herdmodels,deepseekai2025deepseekr1incentivizingreasoningcapability} at layers 3, 12, 19, 25, and 30. For all subject models, features are extracted from either the MLP layers or the residual stream.
As Concept Atlases, we use the Gemma~2~2B model \citep{gemmateam2024gemma2improvingopen} with a Gemma Scope SAE encoder head at layer 20, in both the 16k (\texttt{average\_l0\_71}) and 65k (\texttt{average\_l0\_114}) configurations \citep{lieberum-etal-2024-gemma}. These SAEs are trained on residual stream activations and are referred to as Gemma Scope 16k and Gemma Scope 65k. Unless otherwise specified, Concept Atlas labels are taken from \citet{puri-etal-2025-fade} and generated using automated interpretability methods.

\subsection{Qualitative Results} \label{sec:results-qualitative}

\subsubsection{Identification of Features}

\textbf{Atlas-Alignment} allows us to identify where specific semantic concepts are represented inside a subject model. Starting from a concept of interest in the Concept Atlas, as for example \emph{``chess''}, we map the corresponding Concept Query into the subject model’s latent space and identify the features most strongly aligned with it. Inspecting the top-ranked features and their maximally activating samples can reveal how the specific semantic concept is encoded in the subject model.

Figure \ref{fig:search-qualitative-results} shows examples of Concept Atlas features from Gemma Scope 16k mapped to MLP layers in Llama-Base, Llama-IT, and Llama-R1 using Orthogonal Procrustes. For the shown queries, the highest-scoring subject features activate on samples that are semantically related to the target concepts, demonstrating a flexible and lightweight procedure for detecting concept-relevant features across models without requiring labeled concept datasets.

\subsubsection{Concept Steering}

\textbf{Atlas-Alignment} also enables steering across language models along semantically meaningful directions. Given a concept of interest in the Concept Atlas, we translate the corresponding Concept Query into the subject model’s latent space and use it to bias internal representations during generation, shifting outputs toward the target concept.

\begin{figure*}[t]
    \centering
    \includegraphics[width=\linewidth]{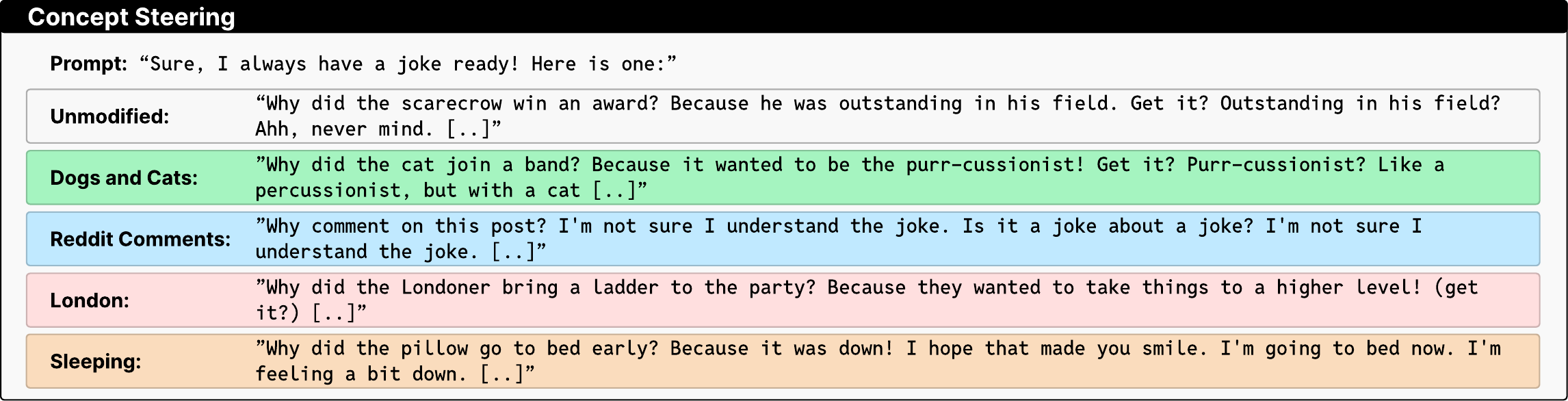}
    \caption{Qualitative concept steering across models. The generation from Llama-IT is steered by injecting translated Concept Atlas directions derived from Gemma Scope 16k using Orthogonal Procrustes. Steering reliably shifts outputs toward the target concepts.}
    \label{fig:steer-qualitative-results}
\end{figure*}

Figure \ref{fig:steer-qualitative-results} shows qualitative examples of concept steering using queries from Gemma Scope 16k mapped to MLP layers of Llama-IT via Orthogonal Procrustes. The resulting generations reflect the intended concepts, demonstrating that Concept Atlas queries alone can guide model outputs. This suggests that \textbf{Atlas-Alignment} transfers not only semantic structure, but also controllability, from the Concept Atlas to the subject model. Additional details are provided in Appendix \ref{sec:appendix:qualitative_steering}.

\subsection{Evaluating Semantic Translations}\label{sec:results-evaluating-translation}

We now turn to a quantitative evaluation of \textbf{Atlas-Alignment}. We focus on the core capabilities the method is designed to support. Specifically, we consider three questions: (1) how well the learned translation preserves semantic structure between the Concept Atlas and the subject model, (2) how effectively it can be used to identify features in the subject model that encode specific concepts, and (3) how well these concepts can be used to steer model behavior. The first two are evaluated in this section, while steering is evaluated in Section \ref{sec:results-evaluating-steering}.

\subsubsection{Translation Quality} \label{sec:results-evaluating-translation-quality}

To measure how well a translation preserves semantic structure between the subject model latent space and the Concept Atlas, we evaluate how consistently it maintains feature-sample relationships. For a given Concept Atlas feature, we rank input samples by how strongly they activate that feature. We then build a query vector containing only this atlas feature, translate it into the subject model space, and rank the samples based on their similarity to the resulting concept vector $s_c$. Comparing the two rankings indicates how well the translation preserves the underlying semantic signal.

We quantify this with the averaged AUROC and Average Precision (AP) metrics. AUROC captures how well the translated feature distinguishes samples that activate the Concept Atlas feature from those that do not, while AP emphasizes precision at the top of the ranking, measuring the degree to which the most salient concept samples remain highly ranked after translation.

We use the test split of 500k sequences and report averages for 100 randomly selected features from the Gemma Scope 16k Concept Atlas. For further details see Appendix \ref{sec:appendix:identification_queries}.

Table \ref{tab:evaluation-search-quality} shows results for MLP layers in GPT-2, Qwen variants and Ministral, as well as multiple layers of Llama-IT. Orthogonal Procrustes consistently achieves the strongest performance across both AUROC and AP. The difference is most pronounced in AP, where Orthogonal Procrustes reaches values of 0.36-0.49 compared to a random baseline of 0.046. Inter-model differences appear to be similar in size to the intra-model scores of the Llama-IT layers. The relative performance of the different alignment methods largely transfers over the latent spaces. Similar patterns are observed for Llama-Base and Llama-R1 (see Appendix \ref{sec:appendix:extended-experimental-results:translation-quality}).

\begin{table*}[t]
\centering
\scriptsize
\begin{tabular}{l*{10}{S}}
\toprule
Latent Space
& \multicolumn{2}{c}{Covariance}
& \multicolumn{2}{c}{Cross Corr.}
& \multicolumn{2}{c}{Linear Reg.}
& \multicolumn{2}{c}{Semantic Lens}
& \multicolumn{2}{c}{Orth. Procrustes} \\
\cmidrule(lr){2-11}
& {AUROC} & {AP}
& {AUROC} & {AP}
& {AUROC} & {AP}
& {AUROC} & {AP}
& {AUROC} & {AP} \\
\midrule
GPT-2
& 0.79 & 0.19 & 0.78 & 0.19 & 0.75 & 0.15 & 0.72 & 0.14 & \bfseries 0.82 & \bfseries 0.36 \\
Qwen 0.5B 
& 0.82 & 0.23 & 0.82 & 0.23 & 0.81 & 0.23 & 0.77 & 0.18 & \bfseries 0.84 & \bfseries 0.39 \\
Qwen 1.5B
& 0.80 & 0.20 & 0.80 & 0.20 & 0.73 & 0.14 & 0.73 & 0.14 & \bfseries 0.84 & \bfseries 0.41 \\
Qwen 3B
& 0.82 & 0.23 & 0.82 & 0.23 & 0.76 & 0.16 & 0.76 & 0.16 & \bfseries 0.83 & \bfseries 0.42 \\
Ministral
& 0.82 & 0.24 & 0.82 & 0.24 & 0.75 & 0.16 & 0.78 & 0.19 & \bfseries 0.86 & \bfseries 0.49 \\
\addlinespace[0.5ex]
\multicolumn{11}{l}{\textbf{Llama-IT}} \\
\addlinespace[0.25ex]
\hspace{1em}Layer 3
& 0.81 & 0.22 & 0.81 & 0.22 & 0.75 & 0.15 & 0.77 & 0.18 & \bfseries 0.83 & \bfseries 0.43 \\
\hspace{1em}Layer 12
& 0.79 & 0.19 & 0.79 & 0.19 & 0.74 & 0.15 & 0.72 & 0.13 & \bfseries 0.81 & \bfseries 0.38 \\
\hspace{1em}Layer 19
& 0.82 & 0.23 & 0.82 & 0.23 & 0.77 & 0.17 & 0.77 & 0.18 & \bfseries 0.86 & \bfseries 0.49 \\
\hspace{1em}Layer 25
& 0.78 & 0.18 & 0.78 & 0.18 & 0.74 & 0.14 & 0.73 & 0.14 & \bfseries 0.83 & \bfseries 0.44 \\
\hspace{1em}Layer 30
& 0.77 & 0.16 & 0.77 & 0.16 & 0.73 & 0.13 & 0.70 & 0.11 & \bfseries 0.83 & \bfseries 0.39 \\
\bottomrule
\end{tabular}
\caption{Translation quality across models and layers. AUROC and AP for aligning subject model latent spaces to the Gemma Scope 16k Concept Atlas. Higher is better; random baselines are AUROC=0.5 and AP=0.046.}
\label{tab:evaluation-search-quality}
\end{table*}

To test the sensitivity of the translation quality on the training dataset size, we subsample the 500k train samples into subsets of 1k, 10k, 50k, 100k and 250k and create translation matrices for the Llama-IT layer 19 subject model and the Gemma-Scope 16k Concept Atlas. We then evaluate the alignment performance on the full 500k-sample test set.
Results are shown in Figure \ref{fig:dataset-ablation-figure} (left side). Orthogonal Procrustes dominates all other methods from 50k onward, with monotonic gains. Linear Regression peaks at 100k before degrading. Interestingly, even at 50k, Orthogonal Procrustes retains most of its performance and is stronger than all other tested methods across all subsets, showing the efficacy of the method and the possibility of cutting compute cost without strong declines in quality. Full results are shown in Table \ref{tab:evaluation-translation-quality-data-ablation-full} in the Appendix.

We also examine the impact of the pooling function used to aggregate token-level activations, comparing max- to mean-pooling. We observe a method-dependent effect on alignment quality: performance varies across methods; however, Orthogonal Procrustes remains the strongest approach under both mean- and max-pooling and achieves its best overall performance with max-pooling. Further details are provided in Appendix \ref{sec:appendix:extended-experimental-results:translation-quality:pooling-variant}.

\subsubsection{Semantic Retrieval} \label{sec:results-evaluating-translation-semantic-retrieval}

We next evaluate how well the learned translations enable the identification of features in the subject model that encode specific concepts. For a given feature $i$ in the subject model, we generate a set of concept-related input sequences {\small$X^{(i)}$}. Averaging their max-pooled activations in the Concept Atlas yields a targeted Concept Query {\small$q_c^{(i)}$}. We map this query into the subject model space via $\Tsc$, and obtain a similarity vector {\small$s_c^{(i)} \in \mathbb{R}^{d_s}$}, which scores all subject model features by their similarity to the concept encoded by feature $i$. A successful retrieval corresponds to ranking the original feature highly, indicating that the translation preserves its semantic identity.

We evaluate retrieval performance using two ranking-based metrics that capture accuracy and confidence. The Mean Reciprocal Rank (MRR) reflects how highly the correct feature is ranked, with a value of one indicating perfect retrieval. The Mean Predicted Probability (MPP) captures confidence by measuring the softmax-normalized probability assigned to the correct feature after z-score normalization of the similarity scores, with a value of one indicating that all probability mass is assigned to the correct feature. Definitions are provided in Appendix \ref{sec:appendix:extended-experimental-results-semantic-retrieval}.

We evaluate retrieval on subject model features from the MLP layer of Llama-IT layer 19, using automatically generated feature descriptions from \citet{choi2024automatic}. Ground-truth concept sets are constructed from synthetic inputs and filtered for reliability (for details see Appendix \ref{sec:appendix:semantic-retrieval-experiment-details}), yielding 454 validated features. Results are reported for Gemma Scope 16k and 65k Concept Atlases.

\begin{table}[h!]
\centering
\scriptsize
\begin{tabular}{l
                S[table-format=1.2] S[table-format=1.2]
                S[table-format=1.2] S[table-format=1.2]}
\toprule
 & \multicolumn{2}{c}{Gemma Scope 16k} & \multicolumn{2}{c}{Gemma Scope 65k} \\
\cmidrule(lr){2-3}\cmidrule(lr){4-5}
\textbf{Method} & {MRR} & {MPP} & {MRR} & {MPP} \\
\midrule
Covariance        & 0.16 & 0.01 & 0.12 & 0.01 \\
Cross Corr.       & 0.25 & 0.01 & 0.26 & 0.02 \\
Linear Reg.       & 0.03 & 0.00 & 0.02 & 0.00 \\
Semantic Lens     & 0.80 & 0.05 & 0.79 & 0.05 \\
Orth. Procrustes  & \textbf{0.97} & \textbf{0.92} & \textbf{0.97} & \textbf{0.94} \\
\bottomrule
\end{tabular}
\caption{Semantic feature retrieval performance. MRR and MPP for retrieving Llama-IT Layer 19 features from Gemma Scope 16k and 65k. Higher is better; random baselines are MRR=0.015 and MPP=0.002.}
\label{tab:evaluation-alignment-result}
\end{table}

Table \ref{tab:evaluation-alignment-result} shows that Orthogonal Procrustes achieves near-perfect retrieval, with MRR and MPP indicating the correct feature is recovered almost every time and with high confidence. Linear regression, covariance, and cross-correlation yield much lower scores, with linear regression close to random baselines. Semantic Lens performs well in MRR but falls behind Orthogonal Procrustes on MPP. The retrieval evaluation shows significant differences between the various translation methods, highlighting the importance of the choice. It also shows that we can reliably retrieve relevant features using the Orthogonal Procrustes translation method.

\begin{figure*}[h]
    \centering
    \includegraphics[width=\textwidth]{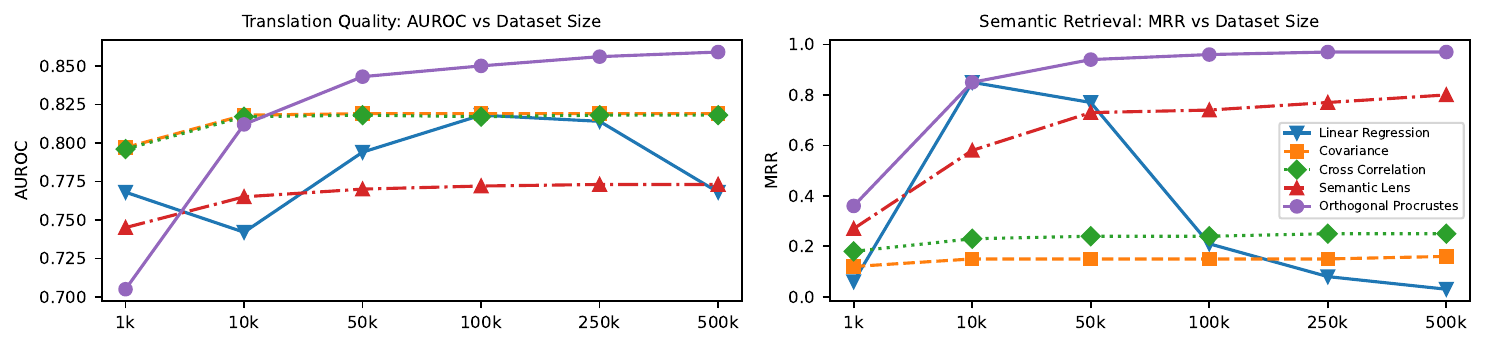}
    \caption{Effect of training data scale on alignment quality. Translation quality (AUROC, left) and semantic retrieval performance (MRR, right) are evaluated on a fixed 500k-sample test set as the training set increases from 1k to 500k samples.}
    \label{fig:dataset-ablation-figure}
\end{figure*}

To measure the sensitivity of semantic retrieval to dataset size we perform the training-data ablation experiment, where the representational alignment methods are used on progressively larger subsampled subsets and tested on the full 500k-sample set. Results are shown in Figure \ref{fig:dataset-ablation-figure} (right side). Orthogonal Procrustes again leads from 50k onward, improving with each larger subset. Linear Regression performs well at the 10k subset, but deteriorates strongly for more samples. Semantic Lens increases monotonically with more training data and Covariance and Cross-Correlation plateau early and stay flat across all subset sizes. Full results are in Table \ref{tab:evaluation-semantic-retrieval-data-ablation-full} in the Appendix.

\subsection{Evaluating Concept-Based Steering}\label{sec:results-evaluating-steering}

\begin{table*}[t]
\centering
\scriptsize
\begin{tabular}{l*{12}{S}}
\toprule
Latent Space
& \multicolumn{2}{c}{Random}
& \multicolumn{2}{c}{Covariance}
& \multicolumn{2}{c}{Cross Corr.}
& \multicolumn{2}{c}{Linear Reg.}
& \multicolumn{2}{c}{Semantic Lens}
& \multicolumn{2}{c}{Orth. Procrustes} \\
\cmidrule(lr){2-13}
& {$f$} & {$\Delta a$}
& {$f$} & {$\Delta a$}
& {$f$} & {$\Delta a$}
& {$f$} & {$\Delta a$}
& {$f$} & {$\Delta a$}
& {$f$} & {$\Delta a$} \\
\midrule

GPT-2
& 1.35 & 0.02 & 5.08 & 0.08 & 13.35 & 0.26 & 2.84 & 0.03 & 1.94 & 0.03 & \bfseries 24.23 & \bfseries 0.84 \\

Qwen 3B
& 1.56 & 0.04 & 3.80 & 0.07 & 10.50 & 0.23 & 2.54 & 0.04 & 2.52 & 0.03 & \bfseries 24.60 & \bfseries 1.14 \\

\addlinespace[0.5ex]
\multicolumn{13}{l}{\textbf{Llama-IT}} \\
\addlinespace[0.25ex]

\hspace{1em}Layer 3
& 1.59 & 0.01 & 3.50 & 0.08 & 2.27 & 0.07 & 2.90 & 0.04 & 3.22 & 0.03 & \bfseries 27.16 & \bfseries 1.23 \\

\hspace{1em}Layer 12
& 5.04 & 0.09 & 39.76 & 1.35 & \bfseries 40.82 & 1.48 & 6.92 & 0.09 & 20.09 & 1.16 & 37.18 & \bfseries 1.93 \\

\hspace{1em}Layer 19
& 3.09 & 0.07 & 19.79 & 0.90 & 30.46 & 1.12 & 5.19 & 0.07 & 7.47 & 0.24 & \bfseries 42.61 & \bfseries 2.38 \\

\hspace{1em}Layer 25
& 2.88 & 0.08 & 10.06 & 0.55 & 16.05 & 0.48 & 3.59 & 0.08 & 5.76 & 0.11 & \bfseries 46.54 & \bfseries 2.32 \\

\hspace{1em}Layer 30
& 7.43 & 0.08 & 19.14 & 0.99 & 32.80 & 1.47 & 5.55 & 0.11 & 8.27 & 0.23 & \bfseries 54.53 & \bfseries 3.04 \\

\addlinespace[0.25ex]
\hspace{1em}Combined
& 1.64 & 0.03 & 36.11 & 1.89 & 48.03 & 2.13 & 7.19 & 0.15 & 23.13 & 0.87 & \bfseries 85.13 & \bfseries 3.91 \\

\bottomrule
\end{tabular}
\caption{Atlas-based steering effectiveness. Faithfulness ($f$) and mean activation change ($\Delta a$) across models and layers. Higher is better; random steering directions serve as a baseline.}
\label{tab:evaluation-steering-result}
\end{table*}

Finally, we evaluate how well different translation methods enable steering a subject model toward specific concepts using only Concept Atlas features. For this we construct queries from Concept Atlas features and map them into the subject model's latent space. The resulting vectors are injected as steering directions, and we measure how strongly they increase the expression of the targeted concept in the generated outputs. 
For a comparison to native steering using dataset-dependent steering vectors computed directly in the subject model, see Appendix \ref{sec:Appendix:Steering_Baseline}.

 We quantify the effectiveness of the steering in two ways: Firstly, we use LLM-based ratings, where the generated sequences are classified as expressing the target concept class or not. We define the faithfulness metric as the maximal relative increase in concept expression over a no-steering baseline, %
 {\small
 \[
\operatorname{f}
= \max_i \frac{r_{\lambda_i} - r_{\lambda_0}}{1 - r_{\lambda_0}} \times 100
\]
}
where $r_{\lambda_i}$ is the share of concept-related generations at steering factor $\lambda_i$, and $r_{\lambda_0}$ is the baseline share without steering. Secondly, we measure the change in activations of the selected features in the Concept Atlas. As the Concept Atlas can also be used as concept-classifier, we expect that sequences generated through successful steering should excite the Atlas features that were used for the creation of the concept-query. We report the strongest average change in Concept Atlas feature activations %
{\small
 \[
\operatorname{\Delta a} 
= \max_i \ \bar{a}_{\lambda_i} - \bar{a}_{\lambda_0}
\]
}%
where $a_{\lambda_i}$ is the Concept Atlas activation at modification factor $\lambda_i$, averaged over the sequence length and the relevant features. We report both metrics averaged across all Concept Queries, to obtain a single value per latent space and translation method.

We construct 10 Concept Queries from Gemma Scope 16k and measure steering effectiveness in the residual streams of GPT-2, Qwen 3B as well as multiple layers of Llama-IT. Each query is evaluated on 32 seed prompts with 50-token continuations. As a baseline, we compare to random steering directions. Further details are provided in Appendix~\ref{sec:appendix:steering_experiment}.

Table \ref{tab:evaluation-steering-result} summarizes steering effectiveness across models and layers. Orthogonal Procrustes most consistently achieves high faithfulness scores, exceeding 24 for GPT-2 and Qwen 3B and reaching up to 85 when applying steering at multiple Llama-IT layers simultaneously. Its mean activation changes are about one-to-two orders of magnitude larger than the random baseline. Covariance and Cross-Correlation methods show strong performance for some layers, particularly in mid-to-late Llama layers, but their effectiveness varies substantially across models and layers. Overall, these results indicate that atlas-based steering can reliably modify model behavior, with Orthogonal Procrustes providing the most consistent and robust approach for semantically controllable interventions.

\section{Conclusion}\label{sec:conclusion} 

In this work, we introduce \textbf{Atlas-Alignment}, a framework for transferring interpretability across language models by aligning unknown latent spaces to a well-understood Concept Atlas. Through quantitative and qualitative evaluations across a range of model families and layers, we show that lightweight linear alignment methods, particularly Orthogonal Procrustes, enable robust semantic transfer, reliable concept retrieval, and controllable generation based solely on Concept Atlas features. These results suggest that interpretability costs can be amortized: a single high-quality atlas can make many downstream models more transparent and controllable at minimal marginal cost.

Our work highlights several directions for future research. While our approach requires only empirically verifiable alignment between a subject model and the chosen Concept Atlas, a broader understanding of when and how widely the Platonic and Linear Representation Hypotheses hold remains an important direction for future work. In addition, characterizing the data requirements of alignment is a promising avenue: although our experiments show that translation quality and semantic retrieval remain strong even with substantially fewer training datapoints, a systematic analysis of these trade-offs would be valuable. Finally, while this work focuses on MLP and residual stream features due to the availability of high-quality labeled latent spaces, a particularly promising extension of \textbf{Atlas-Alignment} is to attention heads, which often implement specialized computations such as context retrieval in Transformer models \citep{kahardipraja2025atlasincontextlearningattention}.

We hope that our approach can serve as a useful basis for future investigations into cross-model alignment as a foundation for interpretability. More broadly, our goal is to make interpretability more scalable and effective by allowing a single high-quality atlas to be used across many different models.

\section*{Acknowledgements}

We sincerely thank Patrick Kahardipraja for his valuable discussions and support. This work was supported by the Federal Ministry of Research, Technology and Space
(BMFTR) as grant xJuRAG (16IS25015B); the German Research Foundation (DFG) as research
unit DeSBi [KI-FOR 5363] (459422098); the Federal Ministry of Education and Research (BMBF) as grant BIFOLD (01IS18025A, 01IS180371I) and the European Union’s Horizon Europe research and innovation programme (EU Horizon Europe) as grant ACHILLES (101189689).

\bibliography{iclr2026_conference}
\bibliographystyle{iclr2026_conference}

\appendix
\onecolumn
\section{Appendix} \label{sec:appendix}

\subsection{Licenses} \label{sec:appendix:licenses}
\texttt{Gemma-2-2b} is released under a custom Gemma Terms of Use. \texttt{Gemma Scope} SAEs are released under Creative Commons Attribution 4.0 International. \texttt{Llama3.1-8B} and \texttt{Llama3.1-8B-Instruct} are released under a custom Llama 3.1 Community License. \texttt{Qwen2.5 0.5B} and \texttt{Qwen2.5 1.5B} are released under the Apache 2.0 License. \texttt{Qwen2.5 3B} is released under a custom Qwen Research License. \texttt{Ministral-8B-Instruct-2410} is released under a custom Mistral AI Research License. The Pile Uncopyrighted dataset, \texttt{GPT-2} and \texttt{Deepseek R1-Distill-Llama-8B} are released under the MIT License.

\subsection{Dataset Generation} \label{sec:appendix:dataset-generation}
For our experiments, we use a subset of the uncopyrighted version of the Pile dataset \citep{gao2020pile}, with all copyrighted content removed. Following the procedure outlined in \cite{puri-etal-2025-fade}, we sample from the test partition while preserving the relative proportions of the original data sources.
The sampled texts are further preprocessed using the NLTK sentence tokenizer \citep{bird2009natural} to divide larger passages into smaller sequences. We then filter out sentences in the bottom and top fifth percentiles of length, which typically correspond to out-of-distribution cases such as single words, isolated characters, or unusually long outliers. We remove sentences consisting only of numbers or special characters and deduplicate the resulting set. The final dataset contains 1M sequences, averaging 120.4 characters with lengths ranging from 2 to 391. Using the Llama 3.1 8B tokenizer, these correspond to an average of 30.1 tokens. We split the data into two subsets of 500k samples each for training and evaluation.

\subsection{Translation Methods} \label{sec:appendix-method-translation-methods}
We list here further translation methods used in this work:

\begin{itemize}
    
\item \textbf{Covariance:}
{\small
\begin{align}
\operatorname{translate}_{\text{ Cov}}(A_s, A_c) = \tilde{A}_s^{\top} \tilde{A}_c \quad \text{with} \quad \tilde{A}_i = A_i - \mu_i
\end{align}
}
where $\mu_i$ is the vector of column means of $A_i$.

\item \textbf{Correlation:}
{\small
\begin{align}
\operatorname{translate}_{\text{Corr}}(A_s, A_c) 
= D_s^{-1}\,\tilde{A}_s^{\top}\tilde{A}_c\,D_c^{-1} 
\quad \text{with} \quad 
\tilde{A}_i = A_i - \mu_i,\;\;
D_i = \operatorname{diag}\!\big(\sigma_i\big)
\end{align}
}
where $\mu_i$ is the vector of column means of $A_i$ and $\sigma_i$ is the vector of column standard deviations of $A_i$.

\item \textbf{Linear Regression:} 
{\small
\begin{align}
\operatorname{translate}_{\text{ OLS}}(A_s, A_c) = \underset{\Tsc}{\argmin} \ \| A_s - A_c \Tsc^\top \|_F^2
\end{align}
}
\item \textbf{Semantic Lens:} A simplified version of Semantic Lens \citep{dreyer2025mechanisticunderstandingvalidationlarge} represents each subject model feature by the set of most activating samples. Specifically, we keep the top-k activations per feature, binarize them, and average the corresponding Concept Atlas embeddings %
{\small
\begin{align}
\operatorname{translate}_{\text{ SL}}(A_s, A_c) = \tilde{A}_s^{\top} A_c \quad \text{with} \quad \tilde{A}_{s \ [ij]} = \frac{1}{k} \cdot \mathbf{1} \left\{ A_{s \ [ij]} \in \text{TopK}(A_{s \ [:, j]}, k) \right\}
\end{align}
}
where TopK returns the indices of the largest $k$ values for the feature column $A_{s [:, j]}$, thus selecting the most salient samples per subject model feature. 

\end{itemize}

\clearpage

\subsection{Qualitative Steering}\label{sec:appendix:qualitative_steering}

We use a short seed prompt and let the model generate 50-100 tokens, while we steer with different modification factors. We apply steering simultaneously to layers 3, 12, 19, 25 and 30 of Llama-IT. We use queries generated from Concept Atlas features from the Gemma Scope 16k SAE in layer 20 and translate them using the Orthogonal Procrustes method. We combine several semantically related features to obtain robust transferable concept directions. All features are weighted equally with a weight of 1. In Table \ref{tab:qualitative_steering_queries} we present the concept name, the used Concept Atlas features along with the used modification factor. 

\begin{table}[h!]
\scriptsize
\centering
\begin{tabularx}{\textwidth}{l|X|c}
\toprule
\textbf{Concept} & \textbf{Features} & \ \textbf{Modification Factor} \\
\midrule
reddit comments  &  [1786, 13945, 9829, 9346, 9736, 13851, 7937, 1914, 2402, 3204, 12203, 10075, 1917, 5067] & 50 \\ 
\hline
dogs and cats  &  [6772, 1089, 12082, 13747] & 50 \\ 
\hline
london  &  [5218, 12614] & 35 \\ 
\hline
sleeping  &  [11097, 8421, 10578, 8158, 9839, 13286, 11572, 8483, 5095, 9972, 2162, 13259, 7574, 1408] & 5 \\ 
\bottomrule
\end{tabularx}
\caption{Abstract concepts and their corresponding features from the Gemma Scope 16k Concept Atlas, along with the modification factor used in the qualitative steering examples.}
\label{tab:qualitative_steering_queries}
\end{table}

\subsection{Evaluation of Translation Quality}\label{sec:appendix:extended-experimental-results}

\subsubsection{Identification Queries}\label{sec:appendix:identification_queries}

We use the following randomly picked 100 Concept Atlas features from the Gemma Scope 16k Concept Atlas to evaluate the strength of ranking capabilities: 

{\scriptsize
\texttt{[464, 470, 496, 648, 662, 708, 775, 837, 908, 1029, 1031, 1217, 1287, 1375, 1554, 1555, 1768, 1796, 1814, 1837, 2385, 2423, 2483, 2712, 2717, 2720, 2782, 2985, 3052, 3258, 4928, 5086, 5219, 5271, 5485, 5544, 5908, 5986, 5992, 6226, 6270, 6371, 6419, 6441, 6525, 6770, 6902, 6930, 7082, 7107, 7190, 7215, 7230, 7291, 7384, 7414, 7647, 7877, 8030, 8332, 8346, 8377, 8391, 8438, 8489, 8598, 8779, 9453, 9622, 9680, 9703, 9743, 9785, 10158, 10242, 10428, 10793, 10819, 10964, 11044, 11200, 11318, 11463, 11668, 11769, 12116, 12373, 12417, 12516, 12549, 12744, 13215, 13293, 13437, 13547, 13708, 13719, 14131, 14660, 14694]}
}

\subsubsection{Full Results of Translation Quality Evaluation}\label{sec:appendix:extended-experimental-results:translation-quality}

In Table \ref{tab:evaluation-search-quality-full} we present the full Results of the Translation Quality Evaluation for the models Llama-Base, Llama-IT, Llama-R1, GPT-2, Qwen 0.5B, Qwen 1.5B, Qwen 3B and Ministral and the Gemma Scope 16k Concept Atlas.

\begin{table*}[h!]
\centering
\scriptsize
\begin{tabular}{l
                S[table-format=1.2] S[table-format=1.2]
                S[table-format=1.2] S[table-format=1.2]
                S[table-format=1.2] S[table-format=1.2]
                S[table-format=1.2] S[table-format=1.2]
                S[table-format=1.2] S[table-format=1.2]}
\toprule
 & \multicolumn{2}{c}{LinReg} 
 & \multicolumn{2}{c}{Cov} 
 & \multicolumn{2}{c}{CrossCorr} 
 & \multicolumn{2}{c}{SemLen} 
 & \multicolumn{2}{c}{OrthProc} \\
\cmidrule(lr){2-3}\cmidrule(lr){4-5}\cmidrule(lr){6-7}\cmidrule(lr){8-9}\cmidrule(lr){10-11}
 & {AUROC} & {AP} 
 & {AUROC} & {AP} 
 & {AUROC} & {AP} 
 & {AUROC} & {AP} 
 & {AUROC} & {AP} \\
 \midrule
\textbf{GPT-2}      & 0.75 & 0.15 & 0.79 & 0.19 & 0.78 & 0.19 & 0.72 & 0.14 & \textbf{0.82} & \textbf{0.36} \\
\textbf{Qwen 0.5B}  & 0.81 & 0.23 & 0.82 & 0.23 & 0.82 & 0.23 & 0.77 & 0.18 & \textbf{0.84} & \textbf{0.39} \\
\textbf{Qwen 1.5B}  & 0.73 & 0.14 & 0.80 & 0.20 & 0.80 & 0.20 & 0.73 & 0.14 & \textbf{0.84} & \textbf{0.41} \\
\textbf{Qwen 3B}    & 0.76 & 0.16 & 0.82 & 0.23 & 0.82 & 0.23 & 0.76 & 0.16 &\textbf{0.83} & \textbf{0.42} \\
\textbf{Ministral}  & 0.75 & 0.16 & 0.82 & 0.24 & 0.82 & 0.24 & 0.78 & 0.19 & \textbf{0.86} & \textbf{0.49} \\
\midrule
\textbf{Llama-Base} \\
L3  & 0.75 & 0.15 & 0.81 & 0.23 & 0.81 & 0.22 & 0.77 & 0.19 & \textbf{0.83} & \textbf{0.43} \\
L12 & 0.75 & 0.15 & 0.79 & 0.19 & 0.79 & 0.19 & 0.73 & 0.13 & \textbf{0.82} & \textbf{0.39} \\
L19 & 0.77 & 0.17 & 0.82 & 0.24 & 0.82 & 0.24 & 0.78 & 0.19 & \textbf{0.86} & \textbf{0.49} \\
L25 & 0.74 & 0.14 & 0.78 & 0.18 & 0.78 & 0.18 & 0.74 & 0.14 & \textbf{0.83} & \textbf{0.44} \\
L30 & 0.74 & 0.14 & 0.78 & 0.16 & 0.78 & 0.16 & 0.71 & 0.11 & \textbf{0.83} & \textbf{0.40} \\
\midrule
\textbf{Llama-IT} \\
L3  & 0.75 & 0.15 & 0.81 & 0.22 & 0.81 & 0.22 & 0.77 & 0.18 & \textbf{0.83} & \textbf{0.43} \\
L12 & 0.74 & 0.15 & 0.79 & 0.19 & 0.79 & 0.19 & 0.72 & 0.13 & \textbf{0.81} & \textbf{0.38} \\
L19 & 0.77 & 0.17 & 0.82 & 0.23 & 0.82 & 0.23 & 0.77 & 0.18 & \textbf{0.86} & \textbf{0.49} \\
L25 & 0.74 & 0.14 & 0.78 & 0.18 & 0.78 & 0.18 & 0.73 & 0.14 & \textbf{0.83} & \textbf{0.44} \\
L30 & 0.73 & 0.13 & 0.77 & 0.16 & 0.77 & 0.16 & 0.70 & 0.11 & \textbf{0.83} & \textbf{0.39} \\
\midrule
\textbf{Llama-R1} \\
L3  & 0.73 & 0.14 & 0.80 & 0.21 & 0.79 & 0.21 & 0.76 & 0.17 & \textbf{0.82} & \textbf{0.40} \\
L12 & 0.73 & 0.13 & 0.78 & 0.17 & 0.77 & 0.17 & 0.71 & 0.12 & \textbf{0.80} & \textbf{0.36} \\
L19 & 0.75 & 0.16 & 0.80 & 0.22 & 0.80 & 0.22 & 0.75 & 0.17 & \textbf{0.84} & \textbf{0.47} \\
L25 & 0.72 & 0.13 & 0.76 & 0.17 & 0.76 & 0.17 & 0.72 & 0.13 & \textbf{0.82} & \textbf{0.42} \\
L30 & 0.73 & 0.13 & 0.76 & 0.15 & 0.76 & 0.15 & 0.70 & 0.10 & \textbf{0.81} & \textbf{0.37} \\
\bottomrule
\end{tabular}
\caption{Full results for the Translation Quality evaluation for single-feature queries from the Gemma Scope 16k Concept Atlas across GPT-2, Qwen 0.5B, Qwen 1.5B and Qwen 3B, Ministral as well as the Llama-Base, Llama-IT, and Llama-R1 models at different layers. Reported are AUROC and AP.  A random AUROC corresponds to 0.5, a random AP to 0.046. Orthogonal Procrustes consistently yields the highest scores across all latent spaces.}
\label{tab:evaluation-search-quality-full}
\end{table*}

\subsubsection{Translation Quality Dataset Size Ablation}\label{sec:appendix:extended-experimental-results:translation-quality:dataset-ablation}

In Table \ref{tab:evaluation-translation-quality-data-ablation-full} we present the full results of the Dataset Size Ablation Experiment for Llama-IT layer 19 and Concept Atlas Gemma Scope 16k.

\begin{table*}[h!]
\centering
\scriptsize
\begin{tabular}{l
                S[table-format=1.2] S[table-format=1.2]
                S[table-format=1.2] S[table-format=1.2]
                S[table-format=1.2] S[table-format=1.2]
                S[table-format=1.2] S[table-format=1.2]
                S[table-format=1.2] S[table-format=1.2]}
\toprule
 & \multicolumn{2}{c}{LinReg} 
 & \multicolumn{2}{c}{Cov} 
 & \multicolumn{2}{c}{CrossCorr} 
 & \multicolumn{2}{c}{SemLen} 
 & \multicolumn{2}{c}{OrthProc} \\
\cmidrule(lr){2-3}\cmidrule(lr){4-5}\cmidrule(lr){6-7}\cmidrule(lr){8-9}\cmidrule(lr){10-11}
 & {AUROC} & {AP} 
 & {AUROC} & {AP} 
 & {AUROC} & {AP} 
 & {AUROC} & {AP} 
 & {AUROC} & {AP} \\
\midrule
\textbf{1k}      & 0.768 & 0.194 & \bfseries 0.797 & \bfseries  0.217 & \bfseries 0.796 & \bfseries 0.215 & 0.745 & 0.149 & 0.705 & 0.205 \\
\textbf{10k}     & 0.742 & 0.252 & \bfseries 0.818 & 0.234 & \bfseries 0.817 & 0.232 & 0.765 & 0.172 & 0.812 & \bfseries  0.388 \\
\textbf{50k}     & 0.794 & 0.315 & 0.819 & 0.235 & 0.818 & 0.234 & 0.770 & 0.179 & \bfseries 0.843 & \bfseries 0.459 \\
\textbf{100k}    & 0.818 & 0.280 & 0.819 & 0.235 & 0.817 & 0.234 & 0.772 & 0.180 & \bfseries 0.850 & \bfseries 0.475 \\
\textbf{250k}    & 0.814 & 0.242 & 0.819 & 0.235 & 0.818 & 0.234 & 0.773 & 0.181 &\bfseries  0.856 & \bfseries 0.488 \\
\textbf{500k}    & 0.768 & 0.170 & 0.819 & 0.235 & 0.818 & 0.234 & 0.773 & 0.181 &\bfseries  0.859 & \bfseries 0.494 \\
\bottomrule
\end{tabular}
\caption{Translation Quality from Llama-IT Layer 19 to Concept Atlas Gemma Scope 16k at different subsets of the train dataset. A random AUROC corresponds to 0.5, a random AP to 0.046.}
\label{tab:evaluation-translation-quality-data-ablation-full}
\end{table*}

\subsubsection{Translation Quality Pooling-Method Evaluation}\label{sec:appendix:extended-experimental-results:translation-quality:pooling-variant}

We examine the impact of the pooling function used to aggregate token-level activations by comparing mean- and max-pooled translation matrices. We construct mean-pooled translation matrices for the subject models GPT-2, Qwen 3B, and Llama-IT (layer 19), and the Gemma Scope 16k Concept Atlas, and evaluate them against the max-pooled counterparts.

Quantitative results for the pooling comparison are reported in Table \ref{tab:evaluation-translation-quality-pooling-experiment}. The effect of pooling choice appears to be method-dependent. Covariance, Cross Correlation, and Semantic Lens generally benefit from mean-pooling, showing higher AP across latent spaces and small AUROC improvements for GPT-2 and Qwen 3B. In contrast, Linear Regression and Orthogonal Procrustes degrade substantially under mean-pooling. Orthogonal Procrustes, however, remains the strongest method overall, retaining the highest AP in both pooling variants while max-pooling further gives it the best AUROC score across all configurations. This shows that pooling choice can strongly influence alignment quality and should be considered jointly with the translation method selection.

\begin{table*}[h!]
\centering
\scriptsize
\begin{tabular}{l l
                S[table-format=1.2] S[table-format=1.2]
                S[table-format=1.2] S[table-format=1.2]
                S[table-format=1.2] S[table-format=1.2]
                S[table-format=1.2] S[table-format=1.2]
                S[table-format=1.2] S[table-format=1.2]}
\toprule
Model & Pooling 
 & \multicolumn{2}{c}{LinReg} 
 & \multicolumn{2}{c}{Cov} 
 & \multicolumn{2}{c}{CrossCorr} 
 & \multicolumn{2}{c}{SemLen} 
 & \multicolumn{2}{c}{OrthProc} \\
\cmidrule(lr){3-4}\cmidrule(lr){5-6}\cmidrule(lr){7-8}\cmidrule(lr){9-10}\cmidrule(lr){11-12}
 & 
 & {AUROC} & {AP} 
 & {AUROC} & {AP} 
 & {AUROC} & {AP} 
 & {AUROC} & {AP} 
 & {AUROC} & {AP} \\
\midrule
GPT-2 & Mean      & 0.696&	0.131&  0.798&	0.230&	0.797&	0.230&	0.769&	0.217&	0.785&	0.310 \\
GPT-2 & Max       & 0.75&	0.15&	0.79&	0.19&	0.78&	0.19&	0.72&	0.14&	\bfseries 0.82&	\bfseries 0.36 \\
\midrule
Qwen 3B & Mean     & 0.725&	0.148&	0.807&	0.254&	0.807&	0.253&	0.792&	0.260&	0.806&	0.351 \\
Qwen 3B & Max      & 0.758 & 0.160 & 0.817  & 0.228 &	0.815 &	0.227 &	0.758 &	0.157 &	\bfseries 0.831 & \bfseries 0.418 \\

\midrule
Llama-IT (Layer 19) & Mean & 0.714&	0.146&	0.807&	0.273&	0.807&	0.271&	0.760&	0.238&	0.841&	0.443 \\
Llama-IT (Layer 19) & Max  & 0.77&	0.17&	0.82&	0.24&	0.82&	0.23&	0.77&	0.18&	\bfseries 0.86&	\bfseries 0.49 \\
\bottomrule
\end{tabular}
\caption{Comparison of Max- and Mean-Pooling Strategies for the models GPT-2, Qwen 3B and Llama-IT layer 19 and the Gemma Scope 16k Concept Atlas measured using the translation quality evaluation.}
\label{tab:evaluation-translation-quality-pooling-experiment}
\end{table*}

\subsection{Semantic Retrieval}\label{sec:appendix:extended-experimental-results-semantic-retrieval}

\subsubsection{Metric Definitions}

The Mean Reciprocal Rank (MRR) measures how highly the correct feature is ranked 
relative to all others:%
{\small%
\begin{align}
\text{MRR} = \frac{1}{d_s} \sum_{i=1}^{d_s} 
\frac{1}{1 + \sum_{k \neq i} \mathbf{1}\!\left[s^{(i)}_i < s^{(i)}_k\right]}.
\end{align}
}%
A score of 1 indicates perfect retrieval. A random baseline score given a random ranking over $d_s$ features results in an MRR of $H_{d_s} / d_s$, where $H_{d_s}$ is the $d_s$-th harmonic number.

The Mean Predicted Probability (MPP) measures the softmax-normalized probability 
assigned to the correct feature after z-score normalization:
{\small
\begin{align}
\text{MPP} = \frac{1}{d_s} \sum_{i=1}^{d_s} 
\frac{\exp\!\left(\tilde{s}^{(i)}_i\right)}
{\sum_{k=1}^{d_s} \exp\!\left(\tilde{s}^{(i)}_k\right)},
\end{align}
}%
where $\tilde{s}^{(i)}$ is the similarity vector standardized by z-score normalization. A score of 1 indicates perfect retrieval. In a baseline case, where all features get assigned the same similarity, the expected MPP is
$1/d_s$.

\subsubsection{Experiment Details}\label{sec:appendix:semantic-retrieval-experiment-details}

We generate 20 synthetic input sequences per feature to form the ground-truth concept sets, on the basis of which we retrieve the features. We use the OpenAI model \texttt{gpt-4o-mini-2024-07-18} \citep{openai2024gpt4o} to create the concept sequences from the labels generated in \citet{choi2024automatic}, using the prompt outlined in Section \ref{sec:dataset-generation-prompt}. To ensure validity, we (i) only include features with reliable descriptions, here defined as features with a harmonic mean score $> 0.75$ on the activation-based FADE metrics \citep{puri-etal-2025-fade} and (ii) discard features where $X^{(i)}$ does not maximally activate the chosen feature in the subject model latent space. This leaves us with a set of 454 validated features.

\subsubsection{Semantic Retrieval Dataset Ablation}\label{sec:appendix:extended-experimental-results-semantic-retrieval:dataset-ablation}

In Table \ref{tab:evaluation-semantic-retrieval-data-ablation-full} we present the full results of the Dataset Size Ablation Experiment for Llama-IT layer 19 and Concept Atlas Gemma Scope 16k.

\begin{table*}[h!]
\centering
\scriptsize
\begin{tabular}{l
                S[table-format=1.2] S[table-format=1.2]
                S[table-format=1.2] S[table-format=1.2]
                S[table-format=1.2] S[table-format=1.2]
                S[table-format=1.2] S[table-format=1.2]
                S[table-format=1.2] S[table-format=1.2]}
\toprule
& \multicolumn{2}{c}{LinReg} & \multicolumn{2}{c}{Cov} & \multicolumn{2}{c}{CrossCorr} & \multicolumn{2}{c}{SemLen} & \multicolumn{2}{c}{OrthProc} \\
\cmidrule(lr){2-3}\cmidrule(lr){4-5}\cmidrule(lr){6-7}\cmidrule(lr){8-9}\cmidrule(lr){10-11}
& {MRR} & {MPP} & {MRR} & {MPP} & {MRR} & {MPP} & {MRR} & {MPP} & {MRR} & {MPP} \\
\midrule
\textbf{1k}   & 0.06 & 0.01 & 0.12 & 0.01 & 0.18 & 0.01 & 0.27 & 0.01 & \bfseries 0.36 &  \bfseries0.15 \\
\textbf{10k}  & \bfseries 0.85 & 0.54 & 0.15 & 0.01 & 0.23 & 0.01 & 0.58 & 0.03 & \bfseries 0.85 & \bfseries 0.68 \\
\textbf{50k}  & 0.77 & 0.35 & 0.15 & 0.01 & 0.24 & 0.01 & 0.73 & 0.04 & \bfseries 0.94 & \bfseries 0.87 \\
\textbf{100k} & 0.21 & 0.03 & 0.15 & 0.01 & 0.24 & 0.01 & 0.74 & 0.04 & \bfseries 0.96 & \bfseries 0.89 \\
\textbf{250k} & 0.08 & 0.01 & 0.15 & 0.01 & 0.25 & 0.01 & 0.77 & 0.05 & \bfseries 0.97 & \bfseries 0.91 \\
\textbf{500k} & 0.03 & 0.00 & 0.16 & 0.01 & 0.25 & 0.01 & 0.80 & 0.05 & \bfseries 0.97 & \bfseries 0.92 \\
\bottomrule
\end{tabular}
\caption{Semantic Retrieval from Llama-IT Layer 19 via Concept Atlas Gemma Scope 16k at different subsets of the train dataset. A random MRR corresponds to 0.015, a random MPP to 0.002.}
\label{tab:evaluation-semantic-retrieval-data-ablation-full}
\end{table*}

\subsection{Steering Evaluation}\label{sec:appendix:steering_experiment}

\subsubsection{Experiment Details}

We use a short seed prompt and let the model generate 50 tokens, while we steer with different modification factors. We use queries generated from Concept Atlas features from the Gemma Scope 16k SAE in layer 20 and translate them using the respective translation methods. Details on the concept queries can be found in Section \ref{sec:appendix:steering_queries}. 

We use the following steering factors in our experiment: $\lambda \in [0, 1, 2, 3, 4, 5, 10, 15, 20, 50, 100]$.
The following seed sequences are used to initialize model generations:

\begin{promptbox}[linewidth=0.5pt]
\scriptsize
\begin{Verbatim}[breaklines=true]
"Once upon a time",
"I just started",
"The candidate",
"Section 1",
"Documents",
"The capital of",
"Let me tell you a joke",
" ",
"French cuisine",
"do_sample=False",
"Would you like to know",
"It is widely used",
"\\scalebox{0.85}{$\\mathcal{L} =",
"The team struggled",
"Hallo, woher denn",
"When you say",
"  ",
"That game",
"I can see",
"Hmm, this",
"I can relate to",
"Who is",
"I understand the",
"Ugh,",
"What the hell was",
"Hey, did anyone",
"Although",
"Thank you for choosing",
"What are you",
"Oh w",
"How dare you open",
"It was my pleasure"
\end{Verbatim}
\end{promptbox}
\label{sec:starting-prompts}

To automatically rate the generated outputs, we use the OpenAI model \texttt{gpt-4o-mini-2024-07-18} \citep{openai2024gpt4o}. The model is instructed to classify sequences according to whether a given concept is expressed: class 0 means the concept is absent, class 2 means it is clearly present, and class 1 (vague/partial) is excluded to reduce spurious correlations. Each sequence is rated three times with temperature 1 and we use the median label as rating. The rating prompt is:

\begin{promptbox}[linewidth=0.5pt]
\scriptsize
\begin{Verbatim}[breaklines=true]
You are given a TOPIC and a set of sequences. Score each sequence for whether the TOPIC is expressed, 
counting either literal string matches or clear semantic references (either is sufficient).
Scoring: 2 = present. 1 = vague/partial. 0 = not expressed. Score independently per sequence.
Each sequence has a unique ID. Provide your ratings as a Python dictionary with sequence IDs as keys and 
respective ratings as values.
Output only the dictionary - no additional text, comments, or symbols.
Below we show examples of input and outputs for three topics.

Example Input: 'Concept: Germany\n
Sequence ID 0: "German scientists unveiled a new fusion experiment."\n
Sequence ID 1: "Berlin's Brandenburg Gate was illuminated last night."\n
Sequence ID 2: "A Munich-based automaker expanded its EV lineup."\n
Sequence ID 3: "Several European countries adjusted tax policy."\n
Sequence ID 4: "The federal republic that reunified in 1990 tightened export rules."\n
Sequence ID 5: "country = \"Deutschland\""\n
Sequence ID 6: "Mount Fuji is a popular destination for hikers."\n
Sequence ID 7: "Silicon Valley companies are racing ahead in AI."' 
Example Output: '{{\"0\": 2, \"1\": 2, \"2\": 2, \"3\": 0, \"4\": 2, \"5\": 2, \"6\": 0, \"7\": 0}}'

Example Input: 'Concept: Football\n
Sequence ID 0: "The quarterback threw a 60-yard touchdown pass."\n
Sequence ID 1: "The UEFA Champions League final kicks off on Saturday."\n
Sequence ID 2: "topic = \"football\""\n
Sequence ID 3: "#football fans filled the stadium after the derby."\n
Sequence ID 4: "The offside rule was explained by the referee."\n
Sequence ID 5: "The match ended 2-1 after extra time."\n
Sequence ID 6: "The chef prepared sushi with fresh tuna."\n
Sequence ID 7: "Quantum entanglement was demonstrated in a new experiment."' 
Example Output: '{{\"0\": 2, \"1\": 2, \"2\": 2, \"3\": 2, \"4\": 2, \"5\": 1, \"6\": 0, \"7\": 0}}'

Example Input: 'Concept: Paris\n
Sequence ID 0: ", there was a young Parisian named Hugo. He found a book filled with maps."\n
Sequence ID 1: "Paris, Texas, USA, 2005, 2006, 2007, 2008, 2009, 2010, 2011, 2012, 2013, 2014, 2015, 2016..."\n
Sequence ID 2: "renowned for its cuisine, and the Parisians are proud of their heritage."\n
Sequence ID 3: "# (1)\nimport numpy as np"\n
Sequence ID 4: "to use the `git` command to get the latest version of the `paris` tool, run these steps..."\n
Sequence ID 5: "You can write 'konnichiwa' in the title, but not the word 'Tokyo'."' 
Example Output: '{{\"0\": 2, \"1\": 2, \"2\": 2, \"3\": 0, \"4\": 2, \"5\": 0}}'
\end{Verbatim}
\end{promptbox}
\label{sec:rating-prompt}

\subsubsection{Steering Evaluation Concept Queries}\label{sec:appendix:steering_queries}

We use queries generated from multiple Concept Atlas features from the Gemma Scope 16k Concept Atlas. All features are weighted equally with a weight of 1. In Table \ref{tab:steering_queries} we show the broad concepts and the feature numbers.

\begin{table}[h!]
\tiny
\centering
\begin{tabularx}{\textwidth}{l|X}
\toprule
\textbf{Concept} & \textbf{Features} \\
\midrule
reddit comments  &  [1786, 13945, 9829, 9346, 9736, 13851, 7937, 1914, 2402, 3204, 12203, 10075, 1917, 5067]  \\ 
\hline
dreams and imagination  &  [5095, 6195, 320, 9017, 9273, 7225, 11922, 1974, 5755, 6576, 13207, 7342, 10331, 2104, 12727, 1631, 10669, 14509, 8630]  \\ 
\hline
gardening  &  [6607, 568, 1689, 13279, 5514, 10459, 3138, 9328, 6056, 1676, 12871, 10010, 5680, 7747, 10759, 6369, 9839, 6316, 9125, 10678, 7360, 12587, 5317, 9396, 2725]  \\ 
\hline
science fiction and fantasy  &  [6020, 9273, 8850, 8544, 3088, 6139, 3120, 6561, 2942, 11922, 2777, 8877, 5267, 6990, 2212, 11512, 13710, 1659, 7995, 10004, 779, 12857, 12899, 7365, 8927, 13805, 13602]  \\ 
\hline
eating  &  [13834, 11544, 1351, 2793, 11867, 5898, 7683, 5531, 9027, 1247, 8513, 9750, 11847, 12394, 5838, 13497, 6621, 11491, 184, 8337, 8991, 668, 1538, 14334, 2480, 1632, 8771, 6657, 9125, 6847, 2247, 13567, 6643]  \\ 
\hline
smart devices  &  [6211, 11318, 257, 5110, 8672, 7105, 14663, 12058, 8356, 1341, 13177, 13000, 948, 5950, 5078, 5791, 13434, 8092, 2942, 5833, 8450, 11350, 8124, 14292, 63, 13363, 1446]  \\ 
\hline
driving and cars  &  [856, 6418, 11182, 3178, 6571, 5814, 11212, 11620, 5877, 14326, 7054, 7732, 8038, 10526, 6964, 10870, 14386, 11957, 10196, 11028, 14566, 13018, 11309, 2724, 9791, 3154, 6375, 7224, 1336, 10172, 207, 9032, 6767, 14137]  \\ 
\hline
dogs and cats  &  [6772, 1089, 12082, 13747]  \\ 
\hline
video games  &  [8877, 13641, 12839, 13962, 3016, 12805, 13317, 13596, 13064, 7522, 14643, 10390, 5864, 8026, 67, 3089, 5380, 2003, 211, 9270, 1537, 14751, 12039, 4950, 6538, 14259, 1276, 5979, 2942]  \\ 
\hline
health and well-being  &  [12413, 986, 5090, 13997, 6624, 12235, 668, 2162, 10317, 155, 11583, 12425, 9404, 2963, 11867, 7943, 2009, 1705, 2912, 10078, 13216, 12593, 1462, 10355, 49, 11043, 5522, 8521]  \\ 
\hline
\bottomrule
\end{tabularx}
\caption{Multi Query Features and the corresponding features from the Gemma Scope 16k Concept Atlas used in the steering evaluation.}
\label{tab:steering_queries}
\end{table}

\subsection{Comparison to Native Steering} \label{sec:Appendix:Steering_Baseline}

\subsubsection{Experiment Details}

We compare supervised steering vectors computed directly in the subject model to translated steering vectors to assess how much performance is preserved under translation, providing a reference for cross-model steering via the learned alignment.

For each concept in Table \ref{tab:steering_queries}, we generate a corresponding synthetic dataset using LLMs. We use the prompt described in Section \ref{sec:dataset-generation-prompt} and the OpenAI model \texttt{gpt-4o-mini-2024-07-18} \citep{openai2024gpt4o}. The datasets comprise between 136 and 171 sequences.
Steering vectors are computed by contrasting the mean last-token activation of samples from one concept with those of the remaining concepts. We compute these vectors both in several layers of the subject model and in the Concept Atlas. We then compare steering performance using native steering vectors and steering vectors translated from the Concept Atlas via Atlas-Alignment. We use multiple layers of Llama-IT as subject models and the Gemma Scope 16k Concept Atlas.

Table \ref{tab:evaluation-baseline-steering-result} summarizes the results. As expected, steering vectors computed directly within the subject model achieve the strongest performance in most settings, particularly with respect to faithfulness, as they rely on concept-specific supervision and direct access to the target representation space. Among translation methods, Orthogonal Procrustes performs best across all layers. It achieves the highest faithfulness of all alignment approaches and closely matches native steering, surpassing it in Layer 3 and approaching it in Layers 25 and 30. Averaged across layers, Orthogonal Procrustes trails native steering by 5.6 faithfulness points, indicating that a large fraction of native steering effectiveness is preserved despite the cross-model translation. In contrast, covariance-based and cross-correlation methods outperform random steering in most layers but show a larger gap to native steering, while linear regression and Semantic Lens yield weaker and less consistent improvements. 

Steering vectors computed directly within the subject model appear to be an upper bound in steering effectiveness, as they are derived from concept-specific supervision and have direct access to the target representation space. In contrast, Atlas-based steering translates concept directions learned in a different model and latent space. Despite this, Orthogonal Procrustes preserves much of the effectiveness of native steering, showing that a substantial part of the semantic structure required for steering is shared across models and can be translated through the representational alignment. These results indicate that \textbf{Atlas-Alignment} can enable effective concept steering even when direct supervision in the subject model is unavailable.

\begin{table*}[h!]
\centering
\scriptsize
\setlength{\tabcolsep}{4.4pt}
\begin{tabular}{lSS|*{12}{S}}
\toprule
Llama-IT
& \multicolumn{2}{c}{Native}
& \multicolumn{2}{c}{Random}
& \multicolumn{2}{c}{Covariance}
& \multicolumn{2}{c}{Cross Corr.}
& \multicolumn{2}{c}{Linear Reg.}
& \multicolumn{2}{c}{Sem. Lens}
& \multicolumn{2}{c}{Orth. Proc.} \\
\cmidrule(lr){2-3}
\cmidrule(lr){4-15}
& {$f$} & {$\Delta a$}
& {$f$} & {$\Delta a$}
& {$f$} & {$\Delta a$}
& {$f$} & {$\Delta a$}
& {$f$} & {$\Delta a$}
& {$f$} & {$\Delta a$}
& {$f$} & {$\Delta a$} \\
\midrule

Layer 3
& 16.85 & \bfseries 0.43
& 2.55 & 0.03
& 4.22 & 0.05
& 2.58 & 0.05
& 3.24 & 0.03
& 2.33 & 0.03
& \bfseries 19.62 & 0.36 \\

Layer 12
& \bfseries 48.61 & 1.03
& 2.86 & 0.04
& 15.30 & 0.17
& 21.42 & 0.47
& 4.25 & 0.06
& 10.62 & 0.21
& 38.40 & \bfseries 1.37 \\

Layer 19
& \bfseries 51.10 & \bfseries 2.22
& 1.91 & 0.03
& 8.80 & 0.15
& 16.52 & 0.28
& 6.45 & 0.08
& 4.90 & 0.09
& 36.67 & 1.50 \\

Layer 25
& \bfseries 34.13 & \bfseries 1.65
& 2.59 & 0.03
& 3.88 & 0.08
& 9.51 & 0.21
& 2.55 & 0.06
& 4.18 & 0.07
& 28.41 & 1.05 \\

Layer 30
& \bfseries 38.31 & 2.11
& 2.55 & 0.03
& 5.31 & 0.10
& 7.51 & 0.10
& 6.57 & 0.06
& 3.61 & 0.07
& 37.82 & \bfseries 2.17 \\

\bottomrule
\end{tabular}
\caption{Steering Baseline Result: Faithfulness ($f$) and mean activation change ($\Delta a$) for several alignment methods.}
\label{tab:evaluation-baseline-steering-result}
\end{table*}

\subsubsection{Data Generation Prompt}\label{sec:dataset-generation-prompt}
Prompt used for synthetic data generation:

\begin{promptbox}[linewidth=0.5pt]
\scriptsize
\begin{Verbatim}[breaklines=true]
You are tasked with building a database of sequences that best represent a specific concept. 
To create this, you will generate sequences that vary in style, tone, context, length, and structure, while maintaining a clear connection to the concept. 
The concept does not need to be explicitly stated in each sequence, but each should relate meaningfully to it. Be creative and explore different ways to express the concept.

Here are examples of how different concepts might be expressed:

Concept: "German language" — Sequences might include German phrases, or sentences.
Concept: "Start of a Java Function" — Sequences might include Java code snippets defining a function.
Concept: "Irony" — Sequences might include ironic statements or expressions.

Provide your sequences as strings in a Python List format.

Example: ["This is a first example sequence.", "Second example sequence but it is much longer also there are somy typos in it.?"]

Output only the Python List object, without any additional comments, symbols, or extraneous content.
\end{Verbatim}

\end{promptbox}

\subsection{Use of Large Language Models}

We used large language models to polish and refine the text for clarity and style.

\end{document}